# Data Mining Approach for Analyzing Call Center Performance

Marcin Paprzycki, Ajith Abraham, Ruiyuan Guo and Srinivas Mukkamala[*]
Computer Science Department, Oklahoma State University, USA
Computer Science Department, New Mexico Technology, USA
{marcin,aa,grui}@cs.okstate.edu, srinivas@cs.nmt.edu

**Abstract.** The aim of our research was to apply well-known data mining techniques (such as linear neural networks, multi-layered perceptrons, probabilistic neural networks, classification and regression trees, support vector machines and finally a hybrid decision tree – neural network approach) to the problem of predicting the quality of service in call centers; based on the performance data actually collected in a call center of a large insurance company. Our aim was two-fold. First, to compare the performance of models built using the above-mentioned techniques and, second, to analyze the characteristics of the input sensitivity in order to better understand the relationship between the performance evaluation process and the actual performance and in this way help improve the performance of call centers. In this paper we summarize our findings.

## 1 Introduction

The performance of the call center depends on the performance of its customer service representatives (CSRs) and the call handling regulations. Most existing large call centers collect data that is then used to assess and improve the performance of its representatives 12513. Typically, such data includes some form of quality assessment, time management, and business processing aspects 37. While, data mining has been applied to analyze the customer behavior with its main aim to improve the customer satisfaction, there is not much research on mining the data of performance of call center representatives. Therefore, the aim of our research is to fill this gap by applying data mining techniques to the combined performance evaluation results collected from five call centers of a large nationwide insurance company. The remaining parts of this paper are organized as follows. In section 2, we summarize the related research that was uncovered, followed by the short description of different data mining techniques used in our research (Section 3). Section 4 introduces the features of the data used in our study and introduces results of our experiments, including some sensitivity analysis. We briefly summarize our findings in Section 5.

## 2 Summary of related research

As indicated above, we were able to find only results related to mining customer-related data. Some vendors of monitoring system such as eTalk and GartnerGroup built data mining tools into their *monitoring systems*. These tools are intended primarily for non-experts, such as supervisors and managers. They can "mine" the available data by asking "what if" questions 8. In this way it was found, for instance, that call transfers frustrate customers. *Predictive modeling* such as decision-tree or neural network based techniques can be used to predict customer behavior. Quaero LLC used such techniques to cluster customers according to their current and their potential value 10. *Textual data mining* has also been applied in the context of call centers. Busemann et al. classified e-mail request from customers based on shallow text processing and machine learning techniques. Their system was able to correctly respond to e-mails with an accuracy of 73% 11. Next, *audio data mining* has been experimented with. ScanSoft used context-free-grammar to parse the speech and follow by Sequence Package Analysis to caption the text to which data mining is applied. This approach allowed capturing early warning signs of caller frustration 4. Finally, *web usage mining* has been applied to web-based activities of call centers. Techniques utilized here are similar to these used in other cases of web mining 9.

## 3    Data mining techniques used

Data mining is an information extraction activity with a goal of discovering hidden facts contained in data(bases). Using a combination of machine learning, statistical analysis, modeling techniques and database technology, data mining finds patterns and subtle relationships in data and infers rules that allow the prediction of future results. There exist a number of popular data mining techniques.

*Multi-layer perceptron* (MLP) is the most popular neural network architecture. It consists of at least three layers, an input layer of source neurons, at least one hidden layer of computational neurons, and an output layer of computational neuron(s). The input layer accepts inputs and redistributes to all the neurons of the middle layer. The neurons in the middle layer detect the features of input patterns and pass the features to the output layer. The output layer uses the features to determine the output patterns.

*Linear neural networks* (LNN) have just two layers: an input layer and an output layer. Linear models have good performance on linear problems. However, they cannot solve more complex problems. Linear networks can be trained to serve as a base comparison for non-linear problems. Linear model is relatively simple and not many parameters need to be selected by the users. We used the standard pseudo-inverse (SVD) linear optimization algorithm.

*Probabilistic neural networks* (PNN) have been developed for classification problems and utilize kernel-based estimation. They usually have three layers: one input layer, one hidden layer and one output layer. The network "embeds" the training cases into the hidden layer, which has as many neurons as there are training cases. The output layer "combines" the estimates and produces the output.

*Classification and regression trees* (CART) are techniques based on the tree structured binary decisions. Each decision tree has internal and leaf nodes. Leaf nodes represent the final decision or prediction. CART labels each leaf node a unique increasing integer number from left to right starting from 1. All the records in the dataset are assigned an integer. CART creates decision trees to predict categorical dependencies by using both categorical and continuous predictors.

*Support vector machine* (SVM) is a binary learning method 12. It conducts computational learning based on structural risk minimization that finds a hypothesis *h* for which the lowest true error is guaranteed. The true error of *h* is the probability that *h* will make an error for an unseen and randomly selected case. An upper bound of the true error can be used for *h*. Support vector machine finds the hypothesis *h* and minimizes the bound of the true error.

Finally, the above-described techniques can be combined and we have utilized a *hybrid* decision tree – neural network technique depicted in Figure 1. In this case, data is fed into the decision tree first and then the leave node information is obtained and added into the dataset used by the neural network as an additional variable (new attribute). For the neural network we have used the multi-layer perceptron with three layers and backpropagation learning for training. Here, the same training parameters were used as for the CART and the perceptron applied to separately to the problem.

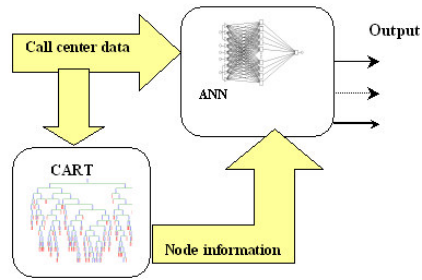

**Fig. 1.** Decision Tree-ANN Hybrid Model

## 4    Call center performance data

The data used in this study is one year worth of actual data from the performance evaluation database of five call centers of a large nationwide insurance company. Here, each customer service representative is being evaluated monthly. To this effect randomly selected calls are recorded (out of ten to sixty calls answered daily by each representative) and the monitoring system constantly keeps up to ten calls for each CSR available. Of these, six randomly selected calls are used by a group of evaluators to assess the CSR's performance. In the insurance company from which the data was obtained, there are two main attributes against which the performance of its representatives is evaluated: (a) *customer service satisfaction* and (b) *business need satisfaction*. The customer service satisfaction score is an *aggregate* result of evaluation based on eleven features. Exactly the same features are used for all products and all call centers. Typical way of evaluating performance with regard to these features is by asking questions like: "did a CSR thanked the customer for calling the company?" or "did a CSR asked what else they can help customer with?" The result of the evaluation is an integer between 0 and 5. Here, 0 means that a given feature was not applicable to the call. A 1 indicates that the CSR did not meet the expectation. A 2 signifies that the expectation was met to some degree (denoted "met some"). A 3 indicates meeting the expectation. A 4 specifies exceeding the expectation. A 5 represents the case when the CSR far exceeded the expectation. These results are then aggregated to a value representing the total level of meeting the customer service satisfaction.

For example, an evaluator reviewed the call and found that only three questions out of eleven were applicable and marked them as 3, 4 and 1 according to how the CSR performed when she/he answered the call. The evaluator also marked the remaining eight questions as 0 (not applicable). The final score of customer service satisfaction was then calculated as the sum (8) divided by the number of applicable questions (3), resulting in the score equal to 2.67. The monthly score is the total score of all applicable questions of all six evaluated calls divided by the total number of applicable questions.

*Business need satisfaction* is scored exactly the same way as the customer service satisfaction. However, the features/questions vary from one product to another. Typical questions are "did a CSR provide correct information to customer" or "did a CSR access proper systems or documents." Depending on the product, the minimum number of the questions is eight and the maximum is sixteen. Although the final scores of customer service satisfaction and business need satisfaction are continuous numbers ranging from 1 to 5, in the call centers, which were the source of the data used in the research, these results are converted to monthly evaluations according to the following rules:

**Table 1.** Rules for converting scores into final evaluation

| | |
|---|---|
| *not met* | score < 2 |
| *met some* | score >= 2 and score < 3 |
| *met* | score >= 3 and score < 4 |
| *exceeded* | score >= 4 and score < 4.75 |
| *far exceeded* | score >= 4.75 |

**Table 2.** Dataset Description

| Category | Attribute Name | Data Type | Format | Example |
|---|---|---|---|---|
| | Agent ID | Integer | | 1, 201, etc |
| | Date of Data | Date | mm/01/yyyy | 09/01/2001 |
| | Training | Boolean | 0, 1 | 0 |
| | Product ID | Integer | | 226, 3927 |
| Quality | Customer Service | Category | 1, 2, 3, 4 | 3 |
| | Business Needs | Category | 1, 2, 3, 4, 5 | 4 |
| Time management | After Call Work Time | Integer | 1, 2, 3, | 180 |
| | Adherence | Float | Percentage | 96% |
| | Attendance | Integer | 1, 2, 3, … | 2 |
| | Auxiliary | Float | Percentage | 4% |

In addition to the above, the attributes of *time management* are utilized and they are: *adherence*, *after call work time*, *auxiliary* and *attendance*. The data of time management is collected from phone switches on monthly basis. *Adherence* is the percentage of the length of time a CSR is logged into the phone switch to the length of time he/she is supposed to be logged in. *After call work time* is the average number of seconds that a CSR spends on post-processing data after calls during a given month. *Auxiliary* is the percentage of the length of time a CSR is spending on personal activity to the length of time that a CSR is logged into the phone switch. *Attendance* is a CSR's monthly absence.

Finally, in the available data, there is a Boolean attribute representing the fact that the CSR is / is not in a training period; each record has a time stamp; and there is an attribute representing which product a CSR is servicing. In summary, there are total ten attributes in the dataset utilized in our project and they are summarized in Table 2.

**4.1 Data Cleaning and Preparation**

As follows from the above, values of customer service and business need satisfaction should fall between one and five. We have therefore removed from the dataset all records with data outside of these bounds. The value of time management categories should all be equal to or above zero. The values below zero are not valid and were deleted. The records that had other missing values were also deleted from the dataset. Finally, when preparing the data, we have found that the distribution of the customer service satisfaction attribute was "bad." Only six records fell into the *not met* and thirteen into the *far exceeded* categories. These records were therefore deleted since they were too few to meaningfully participate in training and testing. The majority of the records fell into the *met* class. This class was thus separated into two sub-classes at 3.5. We have then utilized both the "big" met class and the "sub-class division" and compared the performance of models build for both cases. After cleaning, a total of 14671 records were left in the *customer service* dataset (1469, 5965, 5841, 1396 in subcategories, when the "met" class was separated) and 14690 records in the *business need* dataset (63, 3533, 5974, 3610, 1510 in each category).

Different products have different expected values of after call work, adherence and auxiliary categories. For example, 150 seconds may be a short after call work time for one of the products but a long time for another. Thus the after call work time, adherence and auxiliary were scaled to real numbers from the interval (0, 1). Finally, all of the remaining attributes, except date, were scaled similarly. There are eight input attributes in the final dataset, which are agent ID, date, product ID, training, ACW, aux, adherence and attendance (see Table 2). There are two output attributes: customer service satisfaction and business needs satisfaction. To achieve the best performance, a separate model was built for each of the output attributes. There are four (three) possible output values for the customer service satisfaction and five values for the business needs satisfaction. All the algorithms use random sampling. Each experiment is repeated several times. The results from same algorithm very were close so we could make the assumption that the results are representative.

**4.2 Experiment setup**

For the MLP we used one hidden layer. After a trial and error approach by varying the number of neurons from fifty to a hundred-twenty, we finalized the architecture with 113 neurons. There are eight neurons in the input layer since there are eight input attributes. There is one neuron in each model for one output. We used both a single-phase backpropagation based and a two-phase backpropagation (BP) combined with conjugate gradient (CG) training. We used a typical split of 50% data for training, 25% for testing, and the remaining 25% for cross validation. Same datasets were used for the different machine learning algorithms. We used 100 epochs for both backpropagation and conjugate gradient. In the PNN, we used 7337 neurons for training the customer service attribute and 7346 for business needs attribute in the hidden layer. In the CART algorithm, Gini was selected for goodness of fit measurement to achieve the best performance. We used a maximum tree height of 32 that resulted in the best performance. A hybrid decision tree-neural network was constructed as described in Section 3. For SVM's we used several kernels and after a trial and error approach, we used the third degree polynomial kernel, which resulted in its best performance.

**4.3 Analysis of predictive performance**

The performance measure is calculated from the classification accuracy of testing results. The performance result is the sum of total number correct prediction of the "correct" category and the correct prediction of the "wrong" category divided by the total number of testing cases. The performance of a perfect model is 100% for both the "correct" category and the "wrong" category. The models that have accuracy near 100% are "good." A random classifier should exhibit a 50% accuracy. Table 3 shows the performance of each model for predicting customer service satisfaction. The results of the met class are shown in smaller font as a comparison of separated sub-classes. According to the overall results from the confusion matrix, the ranking of the performance of the trained models is CART, PNN, SVM, BP/CG, BP, Hybrid and the LNN. There are no apparent difference among the BP/CG, BP and the hybrid. For example for the Met 1 class, there were 5969 records out of 14671 falling into "correct" category in the dataset and the remaining 8702 records fell into "wrong" category. CART predicted 4443 out of 5969 correctly, which was 74.43% shown as correct prediction of the "correct" class. CART predicted 6124 out of 8702 correctly, which was 70.37% shown in Table 5. Since 25% of the records in the dataset were used for cross validation for the LNN, MLP, PNN, and SVM, which is different from CART (10 fold cross-validation), the base to calculate the accuracy was different from CART, which was 3668. For example for the met 1 class again, 1448 records out of 3668 fell into the "correct" category and the remaining 2220 records fell into the "wrong" category. 873 records out of 1448 were predicted as "correct" correctly, which is 60.29%. 1359 out of 2220 were predicted as "wrong" correctly, which is 61.21% shown in Table 1. Table 1 also shows the accuracy details for customer service satisfaction. The research predicted the met class and also predicted each met sub-class by splitting the met class into two. Usually the prediction of one large class has higher accuracy. However, it is not true for the met class of customer service satisfaction. The performance for one large class is very close to the performance of predicting sub-classes indicating that the big class has more noise. Our research reveals that the scale used for customer service evaluation is incorrect and mixes data without good differentiation. The CSRs in sub-class 1 are more likely to be met-some performers. The CSRs in sub-class two are more likely to be exceeded performers.

**Table 3.** Classification Accuracy of Customer Service Prediction

| Customer Service Skills – Cross Validation | | | | | | | | | | |
|---|---|---|---|---|---|---|---|---|---|---|
| **Class** | | **Case #** | **Linear %** | **BP %** | **CG %** | **BP/CG %** | **PNN %** | **CART %** | **Hybrid %** | **SVM %** |
| Met Some | Correct | 1469 | 68.77 | 66.77 | 60.28 | 68.88 | 0.00 | 90.13 | 66.96 | 0.00 |
| | Wrong | 13202 | 66.67 | 70.71 | 58.91 | 70.68 | 100.0 | 83.08 | 70.47 | 100.0 |
| | Overall | | 68.56 | 70.38 | 59.04 | 70.52 | 90.26 | 91.65 | 70.33 | 89.95 |
| Met 1 | Correct | 5969 | 58.16 | 60.29 | 54.35 | 60.80 | 28.78 | 74.43 | 62.80 | 18.44 |
| | Wrong | 8702 | 60.31 | 61.24 | 54.77 | 60.73 | 86.37 | 70.37 | 58.66 | 90.64 |
| | Overall | | 59.04 | 60.87 | 54.60 | 60.76 | 63.13 | 74.65 | 60.40 | 61.28 |
| Met 2 | Correct | 5841 | 59.40 | 59.15 | 51.25 | 60.12 | 34.63 | 83.79 | 61.07 | 22.79 |
| | Wrong | 8830 | 59.93 | 61.75 | 52.85 | 62.88 | 81.55 | 63.59 | 61.95 | 88.65 |
| | Overall | | 59.72 | 60.73 | 52.22 | 61.79 | 64.93 | 73.85 | 61.60 | 62.54 |
| Met (1 and 2) | Correct | 11810 | 55.77 | 61.29 | 47.46 | 60.87 | 99.79 | 74.69 | 61.88 | 100.0 |
| | Wrong | 2861 | 54.30 | 62.98 | 45.44 | 62.81 | 0.35 | 83.94 | 61.45 | 0.00 |
| | Overall | | 55.49 | 61.62 | 47.07 | 61.25 | 89.57 | 76.50 | 61.61 | 80.30 |
| Exceeded | Correct | 1396 | 65.58 | 67.25 | 50.29 | 68.71 | 0.00 | 91.12 | 65.08 | 0.00 |
| | Wrong | 13275 | 63.32 | 68.51 | 49.14 | 68.72 | 100.0 | 84.12 | 71.43 | 100.00 |
| | Overall | | 65.37 | 68.39 | 49.25 | 68.72 | 90.97 | 82.36 | 70.85 | 50.35 |

**Table 4.** Classification Accuracy of Business Need Prediction

| Business need Satisfaction - Cross Validation | | | | | | | | | | |
|---|---|---|---|---|---|---|---|---|---|---|
| Class | | Case # | Linear % | BP % | CG % | BP/CG % | PNN % | CART % | Hybrid % | SVM % |
| Not met | Correct | 63 | 50.00 | 53.85 | 53.85 | 53.85 | 0.00 | 100.00 | 65.00 | 0.00 |
| | Wrong | 14608 | 74.80 | 80.24 | 65.70 | 81.91 | 99.97 | 96.45 | 87.92 | 100.00 |
| | Overall | | 74.73 | 80.15 | 65.66 | 81.81 | 99.46 | 99.62 | 87.80 | 99.73 |
| Met some | Correct | 3533 | 76.63 | 80.29 | 43.24 | 79.05 | 52.77 | 93.43 | 91.32 | 57.96 |
| | Wrong | 11138 | 75.33 | 81.14 | 40.66 | 81.90 | 91.73 | 83.38 | 82.63 | 90.59 |
| | Overall | | 76.33 | 80.94 | 41.29 | 81.21 | 82.52 | 89.14 | 82.33 | 82.98 |
| Met | Correct | 5974 | 66.14 | 70.36 | 62.35 | 70.23 | 52.20 | 82.64 | 71.02 | 50.79 |
| | Wrong | 8697 | 60.30 | 68.03 | 59.38 | 67.94 | 81.40 | 75.03 | 69.07 | 90.59 |
| | Overall | | 62.67 | 68.98 | 60.59 | 68.87 | 69.53 | 79.82 | 69.88 | 69.84 |
| Exceeded | Correct | 3610 | 68.31 | 73.77 | 55.77 | 74.22 | 23.10 | 93.82 | 76.52 | 24.57 |
| | Wrong | 11061 | 72.46 | 74.78 | 50.09 | 75.77 | 94.23 | 79.71 | 73.93 | 94.74 |
| | Overall | | 71.46 | 74.54 | 51.53 | 75.38 | 77.12 | 86.51 | 74.59 | 76.75 |
| Far exceeded | Correct | 1510 | 71.03 | 74.92 | 59.22 | 75.83 | 2.12 | 96.82 | 78.00 | 0.00 |
| | Wrong | 13161 | 75.68 | 78.78 | 58.70 | 79.32 | 99.46 | 85.81 | 82.73 | 100.00 |
| | Overall | | 75.17 | 78.43 | 58.74 | 79.00 | 90.69 | 92.33 | 80.84 | 96.12 |

Table 4 shows the performance of each model for predicting business need satisfaction. The way to calculate the performance of business need prediction is exactly the same as the way for customer service. The ranking of the performance is the same as the models for customer service. After looking into the performance accuracy of each correct/wrong class, the research found that PNN models are not valid for the dataset used. The performance of BP/CG is a bit better than BP. However the results are very close and it is not proper to make the conclusion that the models trained by BP/CG have better performance than the ones trained by BP alone. The performance of hybrid model was at least the same as CART. However, the overall accuracy is a bit better than BP and BP/CG models. The LNN model serves as a comparison for other models. The models trained by other algorithms are supposed to have at least the performance that linear models can get. CART models have the best performance in the research. They not only have the best overall performance, but also they have highest accuracy to predict "correct" (C1) and "wrong" (C0) for all each class.

**Table 5.** Ranking of the Inputs (importance) for Predicting Customer Service

| Customer Service Satisfaction - Sensitivity Analysis | | | | | | | | | | |
|---|---|---|---|---|---|---|---|---|---|---|
| Class | Algorithm | Agent | Date | Training | Product | ACW | Adherence | Aux | Attendance | Note |
| Met Some | Linear | 7 | 1 | 5 | 3 | 4 | 2 | 8 | 6 | |
| | BP | 3 | 1 | 2 | 4 | 6 | 7 | 8 | 5 | |
| | BP/CG | 8 | 1 | 3 | 2 | 7 | 5 | 6 | 4 | |
| | Hybrid | 4 | 1 | 8 | 3 | 9 | 6 | 5 | 7 | 2 |
| Met 1 | Linear | 3 | 1 | 5 | 4 | 6 | 8 | 2 | 7 | |
| | BP | 2 | 7 | 6 | 1 | 8 | 4 | 3 | 5 | |
| | BP/CG | 2 | 8 | 6 | 1 | 5 | 3 | 7 | 4 | |
| | Hybrid | 2 | 3 | 5 | 1 | 8 | 4 | 6 | 7 | 9 |
| Met 2 | Linear | 2 | 1 | 5 | 7 | 8 | 3 | 4 | 6 | |
| | BP | 8 | 6 | 7 | 1 | 3 | 2 | 5 | 4 | |
| | BP/CG | 4 | 2 | 5 | 1 | 7 | 3 | 8 | 6 | |
| | Hybrid | 8 | 5 | 9 | 6 | 3 | 2 | 7 | 4 | 1 |
| Met (1 & 2) | Linear | 4 | 1 | 8 | 5 | 2 | 3 | 6 | 7 | |
| | BP | 8 | 1 | 3 | 7 | 6 | 5 | 4 | 2 | |
| | BP/CG | 8 | 1 | 3 | 7 | 6 | 5 | 4 | 2 | |
| | Hybrid | 7 | 1 | 4 | 6 | 3 | 5 | 8 | 9 | 2 |
| Exceeded | Linear | 2 | 1 | 7 | 3 | 5 | 6 | 8 | 4 | |
| | BP | 7 | 1 | 5 | 2 | 4 | 3 | 6 | 8 | |
| | BP/CG | 8 | 1 | 2 | 4 | 5 | 3 | 6 | 7 | |
| | Hybrid | 4 | 3 | 6 | 1 | 9 | 8 | 2 | 7 | 5 |

**4.4. Inputs Sensitivity Analysis**

The sensitivity is calculated by the accumulated errors when a particular attribute is removed from the training. When an attribute is removed from the training model, the higher the error is, the more important the attribute is. The importance of individual inputs is ranked by the accumulated error. Tables 5 and 6 illustrate the ranking of the various attributes for customer service and business needs prediction. First, product is very important to predicting customer service satisfaction, which indicates that CSRs in some products have more opportunity to far exceed than the CSRs in other products. Adherence is important too. Adherence is how much time of the required time a CSR spends logged into the switch and reveals the attitude toward work. A good attitude may lead to good customer service performance. Another interesting characteristic is that date is important when predicting customer service satisfaction. The reason why date is important may be that dates are interrelated with call types. One type of calls may be dominant of all types of calls during a certain period. After that period, calls of another type become the majority in the call volume in next period. Since we are not concerned about the call types in this research (no data is available to mine) we can only speculate that the affect of call types may materialize as the date parameter. Another way to explain the importance of the date may be the training or coaching delivery date. The customer service satisfaction may be improved right after the coaching or training session and may drop after a certain time afterwards. The ranking analysis from the LNN, BP, BP/CG and Hybrid model are pretty consistent in predicting business need satisfaction. The product becomes more important in predicting business needs satisfaction from not met class to the far-exceeded class. This can be interpreted that a CSR has more opportunity to be far exceeding if a CSR services a particular product and less opportunity if he/she services some other product. Agent is more important when predicting exceeded and far-exceeded classes. It means that the top performers are likely staying on the top most of the time. The performance of the CSRs whose performance falls into met or below met is not stable. However, they are more likely staying in met class or below.

**Table 6.** Ranking of the Inputs (importance) for Predicting Business Needs

| Business Need Requirements - Sensitivity Analysis | | | | | | | | | | |
|---|---|---|---|---|---|---|---|---|---|---|
| Class | Algorithms | Agent | Date | Training | Product | ACW | Adherence | Aux | Attendance | Note |
| Not Met | Linear | 2 | 7 | 1 | 6 | 4 | 8 | 5 | 3 | |
| | BP | 7 | 1 | 2 | 6 | 4 | 8 | 3 | 5 | |
| | BP/CG | 7 | 1 | 6 | 8 | 4 | 5 | 2 | 3 | |
| | Hybrid | 4 | 5 | 9 | 2 | 8 | 7 | 3 | 6 | 1 |
| Met Some | Linear | 6 | 2 | 5 | 3 | 4 | 1 | 8 | 7 | |
| | BP | 4 | 3 | 7 | 1 | 8 | 2 | 5 | 6 | |
| | BP/CG | 4 | 3 | 5 | 1 | 8 | 2 | 6 | 7 | |
| | Hybrid | 4 | 2 | 3 | 1 | 8 | 9 | 5 | 6 | 7 |
| Met | Linear | 2 | 1 | 6 | 4 | 3 | 5 | 7 | 8 | |
| | BP | 7 | 1 | 4 | 2 | 8 | 3 | 5 | 6 | |
| | BP/CG | 8 | 1 | 6 | 2 | 7 | 3 | 5 | 4 | |
| | Hybrid | 7 | 3 | 6 | 2 | 8 | 4 | 5 | 9 | 1 |
| Exceeded | Linear | 6 | 8 | 4 | 3 | 2 | 1 | 5 | 7 | |
| | BP | 3 | 6 | 8 | 2 | 4 | 1 | 5 | 7 | |
| | BP/CG | 5 | 3 | 7 | 2 | 4 | 1 | 6 | 8 | |
| | Hybrid | 5 | 2 | 8 | 9 | 4 | 1 | 3 | 7 | 6 |
| Far exceeded | Linear | 2 | 3 | 7 | 1 | 6 | 5 | 4 | 8 | |
| | BP | 2 | 4 | 8 | 1 | 7 | 3 | 5 | 6 | |
| | BP/CG | 2 | 4 | 5 | 1 | 8 | 3 | 6 | 7 | |

| | Hybrid | 3 | 4 | 7 | 1 | 8 | 6 | 5 | 9 | 2 |

## 5 Conclusions

The research built six AI models to predict the quality score of customer service satisfaction and business need satisfaction by using LNN, MLP, PNN, CART, Decision tree-ANN Hybrid model and SVM. The research compared the performance of the six types of models based on the confusion matrix results of cross validation. The performance is also analyzed by using the accuracy of the "correct" category prediction and the accuracy of the "wrong" category prediction. The overall accuracy from CART is 80.63% on predicting customer service satisfaction and 89.48% on predicting business need satisfaction. The accuracy of the "correct" category and the accuracy of the "wrong" category are very close. The trained models based on CART can be used for future prediction. MLP training using BP and CG did not have significant better performance than BP alone. The research also analyzed the sensitivity of inputs. The research found that products, agents and dates could affect the quality of performance more than time management. The CSRs serving in some products have more opportunity to exceed the expectation than the ones in some other products. The top performers constantly exceed or far-exceed the expectation. The performance of CSRs whose evaluation results fall into met or below is not stable. The research suggest that call center management team should focus training and coaching the individuals and products that constantly have low quality instead of emphasizing balancing the length of times spent on calls.

## References

1. A. Gilmore, Call Center Management: Is Service Quality a Priority, Managing Service Quality, vol. 11, no. 3 pp. 153-159.
2. A. Parasuraman, Service Quality and Productivity: a Synergistic Perspective, Managing Service Quality, vol. 12, no. 1, pp. 6-9.
3. Altitude Software, Improving Agent Performance While Maintaining High Level of Motivation, http://www.realmarket.com/required/altitude1.pdf, September 2002, access date: April 2003
4. Amy Neustein, Building Natural Language Intelligence into Voice-Based Applications, http://www.speechtechmag.com/issues/7_4/cover/915-1.html, July 2002, access date: April 2003.
5. David Sims, What Is CRM, http://www.crmguru.com, 2000, access date: April 2003
6. HigherGroud, Discovering the Business Intelligence Hidden in Your Call Center, HigherGroud White Paper Document, June 2002, access date: April 2003.
7. James Brewton, Customer Value-Driven Competency Models: Powerful Tools for Maximizing Call Center Performance and Customer Loyalty, http://www.crm2day.com, 2002, access date: April 2003.
8. Linda Dilauro, What's Next in Monitoring Technology? Data Mining Finds a Calling in Call Centers, Dictaphone Corporation.
9. Pan-Ning Tan, Vipin Kumar, Mining Indirect Associations in Web Data, WebKDD 2001: Mining Log Data Across All Customer Touch Points.
10. Ro King, Data Mining and CRM, http://www.crm2day.com, 2002, access date: April 2002
11. Stephan Busemann, Sven Schmeier, Roman G.Arens, Message Classification in the Call Center, Proceedings of the 6th Conference on Applied Natural Language Processing, Seattle, WA, 2000
12. Vladimir Vapnik, The Nature of Statistical Learning Theory. Springer, New York, 1995.
13. Warren Staples, John Dalrymple, Rhonda Bryar, Accessing Call Center Quality using the SERVQUAL Model, http://www.cmqr.rmit.edu.au/publications/wsjdrbicit02.pdf, 2002, access date: April 2003.